%% file: main.tex
\documentclass{article}

\usepackage{iclr2026_conference,times}
\iclrfinalcopy
\usepackage{times}
\usepackage{latexsym}

\usepackage[table]{xcolor}
\usepackage[most]{tcolorbox}
\usepackage{graphicx}
\usepackage{mathtools}
\usepackage{amssymb}
\usepackage{amsmath}
\usepackage{amsthm}
\usepackage{multirow, booktabs, makecell, adjustbox}
\usepackage{caption}
\usepackage{subcaption}
\usepackage{algorithm}
\usepackage{algpseudocode}
\usepackage{microtype}
\usepackage{hyperref}
\usepackage{url}
\usepackage{tikz}
\usepackage{wrapfig}
\usepackage[T1]{fontenc}
\usetikzlibrary{arrows.meta,positioning}

\newcommand{\rolloutsym}{\csname tau\endcsname}
\newcommand{\method}{\textsc{TRACE}}
\definecolor{turnpositive}{RGB}{0,125,74}
\definecolor{turnnegative}{RGB}{190,45,45}
\definecolor{turnneutral}{RGB}{105,105,105}
\newcommand{\finalcorrect}{%
	{\color{turnpositive}\ensuremath{\checkmark}\ \textbf{Correct}}}
\newcommand{\finalincorrect}{%
	{\color{turnnegative}\ensuremath{\boldsymbol{\times}}\ \textbf{Incorrect}}}
\newenvironment{creditturn}[1]
	{\par\smallskip\begingroup\color{#1}\noindent}
	{\par\endgroup}

\tcbset{
	turncreditexample/.style={
		enhanced,
		breakable,
		colback=white!95!gray,
		colframe=black,
		fonttitle=\bfseries,
		boxrule=0.5pt,
		arc=1mm,
		before skip=0.8em,
		after skip=0.8em
	}
}

\definecolor{darkblue}{rgb}{0, 0, 0.5}
\hypersetup{colorlinks=true, citecolor=darkblue, linkcolor=darkblue, urlcolor=darkblue}

\title{TRACE: Turn-level Reward Assignment via Credit Estimation for Long-Horizon Agents}

\author{
Leitian Tao$^{1,2}$ \quad
Baolin Peng$^2$ \quad
Wenlin Yao$^2$ \quad
Tao Ge$^2$ \quad
Hao Cheng$^2$ \quad
Mike Hang Wang$^2$ \quad\\
\textbf{Jianfeng Gao}$^2$ \quad
\textbf{Sharon Li}$^1$ \\
$^1$University of Wisconsin--Madison \quad
$^2$Microsoft Research
}

\begin{document}

\maketitle

\begin{abstract}
Multi-turn agents solve complex tasks through extended sequences of tool
interactions before producing a final answer, making credit assignment a
fundamental challenge during post-training. Outcome rewards provide reliable
supervision for short-horizon reasoning, but become sparse and high-variance as
trajectories grow to tens or hundreds of tool calls. They can also be
misleading: a failed rollout may contain many useful actions that move the agent
closer to the goal, yet outcome-only training assigns them the same negative
advantage as the eventual mistake. We propose \method{} (Turn-level Reward
Assignment via Credit Estimation), a dense credit-assignment method for agentic
reinforcement learning. \method{} represents rollouts as state transitions at
tool-call boundaries, obtains gold-answer log-probabilities from a frozen
reference model, transforms them into log-ratio state values, and derives
per-action rewards as Temporal-Difference changes in those values. This requires no additional
critic or process-label training, and its one-step log-ratio TD component
telescopes across redundant tool calls. On long-horizon complex search,
\method{} substantially improves base-model tool-use ability using pure RL,
without a cold-start supervised fine-tuning stage, an agentic mid-training
stage, or training on live-web data. On the closed-web BrowseComp-Plus benchmark,
it raises Qwen3-4B from $7.2$ to $35.6$ and Qwen3-30B-A3B from $8.4$ to $42.6$.
The learned search behavior also transfers to open-web benchmarks, and the
learning curves show earlier improvement and faster convergence during RL
training.
\end{abstract}

\input{sections/1_introduction}
\input{sections/2_preliminaries}
\input{sections/3_method}
\input{sections/4_experiments}
\input{sections/5_relatedwork}
\input{sections/6_limitations}
\input{sections/6_conclusion}

\bibliography{dlg-agenticmodeling}

\appendix
\input{sections/7_appendix}

\end{document}

%% file: sections/1_introduction.tex
\section{Introduction}
\label{sec:introduction}

Large language model (LLM) agents increasingly solve complex tasks by reasoning
and acting through many interactions with external environments, including web
navigation, software engineering, and general computer use
\citep{yao2023react, schick2023toolformer, nakano2021webgpt, yao2022webshop,
deng2023mind2web, zhou2024webarena, jimenez2024swebench, xie2024osworld}.
Reinforcement learning with verifiable rewards (RLVR) has been effective for
single-turn reasoning tasks such as mathematics and coding, where a deterministic
checker can evaluate the final answer and provide a clean outcome reward
\citep{shao2024deepseekmath, guo2025deepseekr1}. Applying the same idea to
agents is appealing but insufficient: an agentic rollout may contain dozens of
search, open, find, code, or other tool-use decisions before the final answer is
verified. A single terminal reward says whether the trajectory succeeded, but it
does not reveal which intermediate actions gathered necessary evidence, which
were redundant, or which derailed the agent.

\begin{figure*}[t]
\centering
\includegraphics[width=0.9\linewidth]{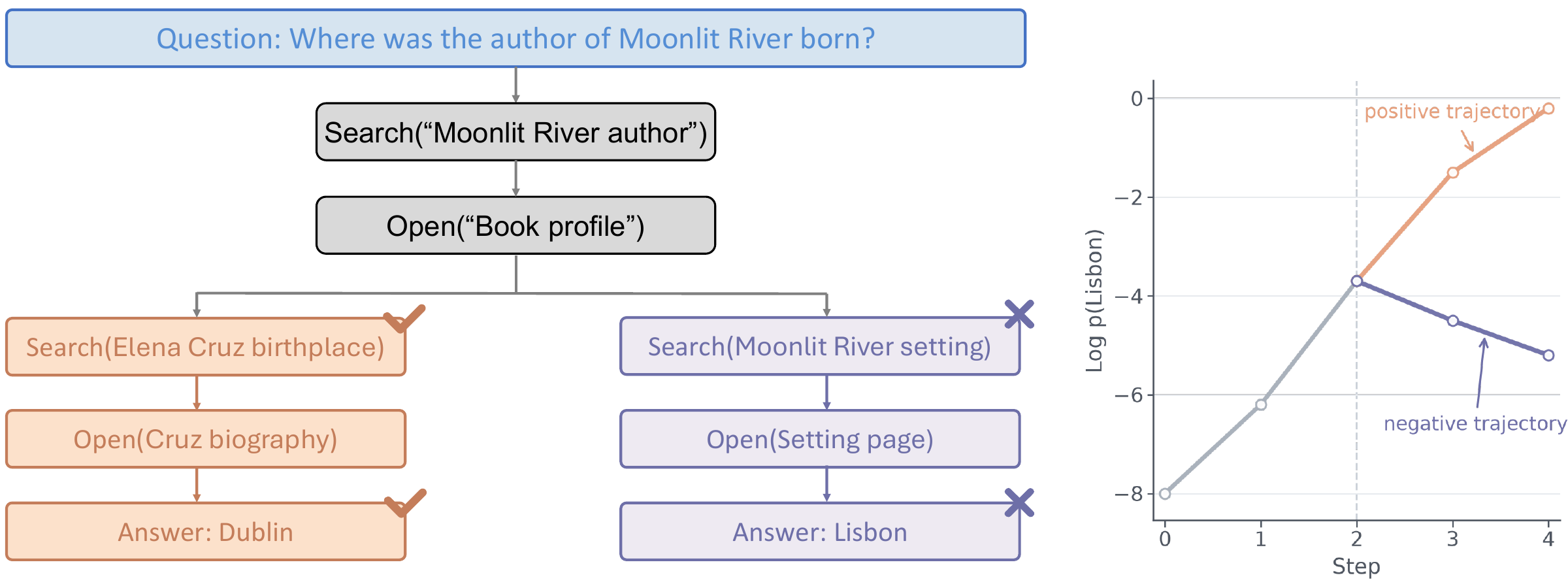}
\caption{\textbf{Credit assignment at tool-call boundaries in a search
trajectory.} The figure illustrates a long-horizon search rollout decomposed
into tool-call turns. Early search and open actions can add task-relevant
evidence to the transcript even when a later branch leads to an incorrect final
answer. Outcome-reward training attaches one trajectory-level advantage to all
actions in the rollout, whereas \method{} computes prefix values at tool
boundaries and assigns turn credit from adjacent value changes.}
\label{fig:teaser}
\end{figure*}

Dense turn-level rewards are therefore desirable, but obtaining them is
challenging because partial progress in agent trajectories is not directly
observed. As illustrated in Figure~\ref{fig:teaser}, a failed rollout can begin
with useful search and open actions that make the answer easier to infer, while
a successful rollout can contain redundant searches or accidental openings that
do not contribute to the final answer. Outcome-only training assigns the same
trajectory-level advantage to all of these turns, under-crediting productive
exploration, over-crediting irrelevant actions, and increasing gradient variance
as horizons grow \citep{sutton2018reinforcement, arjona2019rudder, ye2025beyond}.
Prior work on process supervision offers finer feedback, but typically requires
step-level labels, a strong LLM judge to score intermediate behavior
\citep{zheng2023judging}, or a trained process reward model whose scores may
drift away from final-answer correctness
\citep{uesato2022process,lightman2023verify,wang2024mathshepherd,setlur2024rewarding,yuan2024free}.
This raises the central question of this work: \emph{how can we identify the
tool calls that actually move a long-horizon agent toward the answer, and turn
that signal into dense rewards without relying on step labels, a strong judge,
or a trained process reward model?}

We propose \method{} (\textbf{T}urn-level \textbf{R}eward \textbf{A}ssignment
via \textbf{C}redit \textbf{E}stimation), a critic-free credit-assignment
framework that keeps the final verifier as the anchor while adding dense credit
at tool-call boundaries. The key idea is to use a frozen reference model not as a
judge, but as a stable probe of whether each trajectory prefix makes the gold
answer more predictable. \method{} represents a rollout as state transitions at
tool-call boundaries, scores each prefix by the reference model's gold-answer
log-probability, and converts the score into a log-ratio state value measuring
progress toward the answer. It then assigns turn rewards using
temporal-difference (TD) changes between adjacent values: a tool call receives
positive credit when its returned observation increases answer predictability,
near-zero credit when it adds no useful evidence, and negative credit when it
moves the trajectory away from the answer. Because the one-step TD credits
telescope, redundant intermediate turns cannot inflate this component, while the
cumulative credit remains aligned with the final reference-model state. The
resulting rewards combine this dense TD signal with the standard outcome-level
advantage \citep{sutton1988learning}, preserving verifiable success as the final
training target while distinguishing which turns deserve credit.

We evaluate \method{} on long-horizon complex search, a setting where agents
must learn to interact with an environment, explore through multiple tool calls,
and refine later actions from earlier observations. Because common multi-hop
benchmarks can be solved by strong agents in only a few turns
\citep{yang2018hotpotqa}, we train on deeper synthetic search questions and
evaluate on both closed-web and open-web deep-research benchmarks. Our recipe is
pure RL: it uses outcome-level and turn-level rewards directly, without a
cold-start supervised fine-tuning stage, an agentic mid-training stage, live-web
training data, a strong judging model, or a trained process reward model. Despite
this minimal supervision, \method{} substantially improves base-model tool use
in both closed-web and open-web settings. On closed-web BrowseComp-Plus
\citep{chen2025browsecompplus}, it raises Qwen3-4B from $7.2$ to $35.6$ and
Qwen3-30B-A3B from $8.4$ to $42.6$. The learned behavior also transfers to
open-web retrieval, with the 30B-A3B agent reaching $12.9$ on BrowseComp,
$52.0$ on GAIA, and $45.0$ on xbench-DeepSearch. In addition to final
performance, the training curves show that \method{} starts improving earlier
and converges faster, indicating that turn-level credit makes long-horizon tool
use easier to learn from pure RL. These results show that verifier-anchored
turn-level credit can teach base models to explore, interact with the
environment, and refine tool use for complex long-horizon tasks, while remaining
independent of expensive process supervision or strong judge-based feedback.

%% file: sections/2_preliminaries.tex
\section{Preliminaries}
\label{sec:preliminaries}
\subsection{Agentic Reinforcement Learning}
\label{sec:preliminaries:grpo}

In agentic reinforcement learning, a policy LLM $\pi_\theta$ solves a prompt
$x\sim\mathcal{D}$ by interleaving assistant tokens, tool calls from an
available tool set $\mathcal{T}$, and tool observations before producing a final
answer. A common KL-regularized training objective is
\begin{equation}
\max_{\pi_\theta}
\mathbb{E}_{x \sim \mathcal{D},\, \tau \sim \pi_\theta(\cdot \mid x;\mathcal{T})}
\left[ r_\phi(x,\tau) \right]
-
\beta_{\mathrm{KL}}\,
\mathbb{D}_{\mathrm{KL}}\!\left[
\pi_\theta(\tau \mid x;\mathcal{T})
\,\|\,
\pi_{\mathrm{ref}}(\tau \mid x;\mathcal{T})
\right],
\label{eq:agentic-rl-objective}
\end{equation}
where $\tau$ denotes the complete rollout, $r_\phi$ is the reward function,
$\pi_{\mathrm{ref}}$ is a reference policy, and $\beta_{\mathrm{KL}}$ controls the strength of
the KL penalty. In reinforcement learning with verifiable rewards, $r_\phi$ is
usually an outcome reward that is observed only after the rollout is complete,
for example by checking whether the final answer satisfies an automatically
verifiable criterion \citep{sutton2018reinforcement,shao2024deepseekmath}.
Unlike single-turn reasoning, agentic rollouts induce a distribution over both
the intermediate interaction trajectory and the final answer
\citep{yao2023react,schick2023toolformer,nakano2021webgpt,yao2022webshop}. We
write the interaction part as alternating policy actions and environment
observations. Let $\mathcal{R}=((a_1,o_1),\ldots,(a_{T_{\mathcal{R}}},o_{T_{\mathcal{R}}}))$ and
$H_k=(x,a_1,o_1,\ldots,a_{k-1},o_{k-1})$. Then
\begin{equation}
P_\theta(\mathcal{R}, y \mid x;\mathcal{T}) =
\underbrace{
\prod_{k=1}^{T_{\mathcal{R}}}
\pi_\theta(a_k \mid H_k;\mathcal{T})\,
P_{\mathrm{env}}(o_k \mid H_k,a_k;\mathcal{T})
}_{\mathrm{Tool\ interaction}}
\cdot
\underbrace{
\prod_{t=1}^{T_y}
\pi_\theta(y_t \mid y_{<t}, H_{T_{\mathcal{R}}+1};\mathcal{T})
}_{\mathrm{Answer\ generation}},
\label{eq:tir}
\end{equation}
where $\mathcal{R}$ is the multi-turn trajectory, including tool-call
feedback, and $y$ is the final answer. Gradients are taken through assistant
actions and answer tokens, while tool observations are environment outputs rather
than policy-generated tokens. This decomposition highlights the source of the
credit-assignment problem: the terminal reward supervises a long sequence of
heterogeneous policy decisions, but does not identify which turns made the answer
more or less likely.

For credit assignment, it is useful to view the interaction trajectory at
tool-call boundaries. Given a rollout with $T_{\mathcal{R}}$ tool interactions,
let $S_k$ denote the prefix containing the prompt and the first $k$
action--observation pairs, optionally followed by the final-answer opener used
to score answer tokens. The transition from $S_k$ to $S_{k+1}$ is induced by the
next policy action and environment observation $(a_{k+1},o_{k+1})$. These
prefix transitions are the natural units for assigning credit to tool use,
because they isolate how each interaction changes the information available to
the agent.

Typical RL training builds on outcome-based agentic RL algorithms such as Group
Relative Policy Optimization (GRPO)~\citep{guo2025deepseekr1}, which compare
multiple rollouts from the same prompt and avoid a learned critic for the
outcome-reward stage \citep{shao2024deepseekmath}. However, the resulting
group-relative advantage is still attached to the entire rollout. It can rank a
successful trajectory above failed alternatives, but it cannot distinguish
useful tool calls from redundant or harmful ones within the same trajectory.
This limitation motivates a reward construction that preserves the reliability
of outcome verification while exposing denser signal at the prefix-transition
level.

\subsection{Temporal-Difference Credit Assignment}
\label{sec:preliminaries:td}

Temporal-difference (TD) learning estimates progress by comparing the value of
two adjacent states \citep{sutton1988learning}. Given a value function $V(s)$,
an immediate reward $r_t$, and a discount $\gamma$, the one-step TD error is
\begin{equation}
\delta_t \;=\; r_t + \gamma V(s_{t+1}) - V(s_t).
\label{eq:td-error}
\end{equation}
The TD error is positive when the transition leads to a state whose realized or
estimated future return is higher than expected, and negative when it leads to a
worse state. In sparse-reward agentic tasks, intermediate environment rewards are
usually unavailable. If we set $r_t=0$ before termination and use
$\gamma = 1$, the TD form reduces to the value difference
$V(s_{t+1}) - V(s_t)$, which naturally measures the incremental progress made by
one transition. The difficulty is that long-horizon agent tasks usually lack a
reliable dense value function over partially completed tool trajectories. The
next section addresses this missing ingredient: how to obtain a stable
prefix-level progress signal without training a critic or collecting step-level
process labels.

%% file: sections/3_method.tex
\section{Turn-Level Reward Assignment via Credit Estimation}
\label{sec:method}

\begin{figure}[t]
\centering
\includegraphics[width=\linewidth]{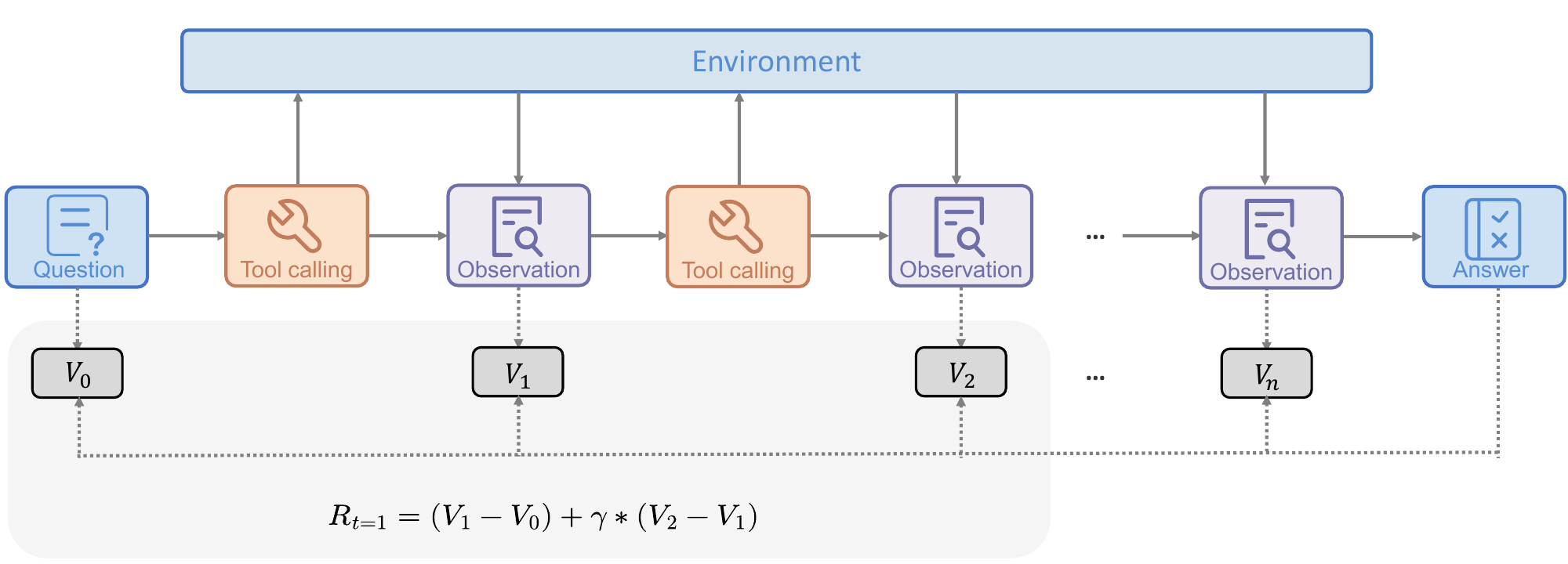}
\caption{\textbf{\method{} reward construction at tool-call boundaries.} The
figure shows a long-horizon rollout segmented by tool calls and observations.
Each prefix state is scored by a frozen reference model, transformed into a
log-ratio value, and compared with adjacent prefix values to obtain turn-level
TD credit. The turn-level advantage is then combined with the trajectory-level
outcome advantage for policy optimization.}
\label{fig:trace-framework}
\vspace{-0.5em}
\end{figure}

We introduce \method{}, a critic-free credit-assignment method for long-horizon
agent post-training. Instead of learning a critic over sparse terminal rewards,
\method{} measures whether each tool interaction makes the gold answer more
predictable under a frozen reference model. It splits rollouts at tool-call
boundaries, maps prefix answer log-probabilities to log-ratio state values, uses
TD changes in those values as turn-level rewards, and mixes the resulting
turn-level advantage with the standard GRPO outcome advantage. The outcome reward
remains the final arbiter of task success. Figure~\ref{fig:trace-framework}
illustrates the framework, and Algorithm~\ref{alg:turn-reward} summarizes the
procedure.
\subsection{Tool-Boundary States}
\label{sec:method:setup}

Following the prefix-transition view in Section~\ref{sec:preliminaries:grpo},
we instantiate the notation used by \method{}. For a sampled trajectory on
question $x$ with gold answer $y^\star$, write
\begin{equation}
\rolloutsym
=\bigl(x,\;a_1,\;o_1,\;a_2,\;o_2,\;\ldots,\;
 a_T,\;o_T,\;\hat{y}\bigr),
\label{eq:rollout}
\end{equation}
where $a_k$ is the $k$-th assistant action, $o_k$ is the corresponding tool
observation, $\hat{y}$ is the final answer span, and $T$ is the number of tool
interactions. The terminal verifier returns $R=R(\hat{y},y^\star)$. \method{}
constructs one credit target for each transition $S_k\to S_{k+1}$ induced by
$(a_{k+1},o_{k+1})$, while the policy-gradient loss is applied only to
policy-generated assistant actions and answer tokens; tool observations remain
environment outputs and are masked from the loss. The remaining question is how
to score whether a prefix transition improves the trajectory's ability to reach
the gold answer.

\subsection{State Value via Log-Ratio Answer Probability}
\label{sec:method:value}

\paragraph{Reference answer score.}
Long-horizon agentic RL makes critic learning brittle: the trajectory reward is
sparse, observations shift the state after each tool action, and failed
trajectories can still contain useful action--observation transitions. \method{}
therefore replaces a learned critic with a \emph{frozen} reference model
$\pi_{\mathrm{ref}}$ (a copy of the policy initialization), used only to estimate
how ready the single-trajectory prefix is to produce the gold answer. This is a
natural proxy for progress in RLVR settings: the gold answer is available during
training, and a transcript that has collected relevant evidence should make that
answer easier for a fixed language model to conditionally predict. Freezing the
model keeps this probe stable across policy updates and prevents it from becoming
another learned reward model. For state
$S_k$, the average gold-answer log-probability is
\begin{equation}
\bar{\ell}_k
=\frac{1}{|y^\star|}\sum_{t=1}^{|y^\star|}
\log \pi_{\mathrm{ref}}\!\left(y^\star_t \mid S_k,\, y^\star_{<t}\right)
\;\le\; 0 .
\label{eq:ref-logprob}
\end{equation}
Larger (less negative) $\bar{\ell}_k$ means that the current transcript already
contains more evidence for generating $y^\star$. All $\bar{\ell}_k$ are obtained
in a single batched forward pass and are never optimized; final-answer
correctness is still determined by the outcome verifier.

\paragraph{Log-ratio state value.}
The score $\bar{\ell}_k$ is a prefix-readiness measure, but its raw scale is not
directly suitable for credit assignment. Raw changes in $\bar{\ell}_k$ can be
misleading: the
same absolute gain is minor when the trajectory is far from the answer but
important when it nearly resolves the answer. For instance, shrinking the
remaining gap from $0.2$ to $0.1$ removes half the uncertainty, whereas
$5.1\!\to\!5.0$ barely changes it despite an identical raw gain. We therefore
model the state value as relative closure of the initial answer-likelihood gap.
Define $d_k=-\bar{\ell}_k+\epsilon$ with offset $\epsilon>0$, and set
\begin{equation}
V(S_k)\;=\;\log\frac{d_0}{d_k}
\;=\;\log\frac{-\bar{\ell}_0+\epsilon}{-\bar{\ell}_k+\epsilon}.
\label{eq:value}
\end{equation}
Thus $V(S_0)=0$, and a larger $V(S_k)$ means that the action--observation history
has closed a larger \emph{fraction} of the initial gap. The later TD difference
$V(S_{k+1})-V(S_k)$ therefore measures the value added by the transition induced
by action $a_{k+1}$ and observation $o_{k+1}$.

\begin{algorithm}[h]
\caption{\method{} reward construction and policy update for one prompt group.}
\label{alg:turn-reward}
\begin{algorithmic}[1]
\Require prompt $x$, gold answer $y^\star$, current policy $\pi_\theta$, behavior snapshot $\pi_{\mathrm{old}}$, frozen reference $\pi_{\mathrm{ref}}$, group size $G$, horizon $K$, discount $\gamma_{\mathrm{td}}$, offset $\epsilon$, terminal scale $\lambda_{\mathrm{term}}$, weights $\alpha_{\mathrm{out}},\alpha_{\mathrm{turn}}$, clip bounds $c_-,c_+$
\State Sample $G$ trajectories $\{\rolloutsym_g\}_{g=1}^{G}\sim\pi_{\mathrm{old}}(\cdot\mid x)$,
where $\rolloutsym_g=(x,a_{g,1},o_{g,1},\ldots,a_{g,T_g},o_{g,T_g},\hat{y}_g)$.
\State Compute terminal rewards $R_g=R(\hat{y}_g,y^\star)$ and outcome advantages
$A^{\mathrm{out}}_g=\operatorname{GroupNorm}(R_g;\{R_h\}_{h=1}^G)$, returning $0$ for all $g$ if the group standard deviation is $0$.
\For{$g=1,\ldots,G$}
  \State Form answer-ready prefixes $S_{g,0},\dots,S_{g,T_g}$ from $\rolloutsym_g$.
  \For{$k=0,\ldots,T_g$}
    \State $\displaystyle \bar{\ell}_{g,k}=\frac{1}{|y^\star|}\sum_{t=1}^{|y^\star|}\log \pi_{\mathrm{ref}}(y^\star_t\mid S_{g,k},y^\star_{<t})$.
    \State $\displaystyle d_{g,k}=-\bar{\ell}_{g,k}+\epsilon,\qquad V_{g,k}=V(S_{g,k})=\log\frac{d_{g,0}}{d_{g,k}}$.
  \EndFor
  \For{$k=0,\ldots,T_g-1$}
    \State $\displaystyle \delta_{g,k}=V_{g,k+1}-V_{g,k}=\log\frac{d_{g,k}}{d_{g,k+1}}$.
    \If{$K=0$}
      \State $c^{(K)}_{g,k}=0$ and $h_{g,k}=k$.
    \Else
      \State $\displaystyle h_{g,k}=\min(k+K-1,T_g-1)$.
      \State $\displaystyle Z_{g,k}=\sum_{u=k}^{h_{g,k}}\gamma_{\mathrm{td}}^{u-k},\qquad
      c^{(K)}_{g,k}=Z_{g,k}^{-1}\sum_{u=k}^{h_{g,k}}\gamma_{\mathrm{td}}^{u-k}\delta_{g,u}$.
    \EndIf
    \Statex \hspace{\algorithmicindent}$\displaystyle
    \begin{aligned}
    r^{\mathrm{turn}}_{g,k}={}&c^{(K)}_{g,k}
    +\mathbf{1}[h_{g,k}=T_g-1]\lambda_{\mathrm{term}}\gamma_{\mathrm{td}}^{T_g-k}A^{\mathrm{out}}_g .
    \end{aligned}$
  \EndFor
\EndFor
\State Let $\mathcal{I}_g$ be the assistant-token indices of trajectory $g$.
\State For each $g$ and $t\in\mathcal{I}_g$, set
$\displaystyle \hat{A}_{g,t}=\alpha_{\mathrm{out}}A^{\mathrm{out}}_g+
\alpha_{\mathrm{turn}}r^{\mathrm{turn}}_{g,\mathrm{turn}(t)}$.
\State $\displaystyle \rho_{g,t}(\theta)=\frac{\pi_\theta(a_{g,t}\mid s_{g,t})}{\pi_{\mathrm{old}}(a_{g,t}\mid s_{g,t})}$.
\State Update $\pi_\theta$ by maximizing $\mathcal{J}_{\mathrm{TRACE}}(\theta)$:
\Statex \hspace{\algorithmicindent}$\displaystyle
\frac{1}{G}\sum_{g=1}^{G}\frac{1}{|\mathcal{I}_g|}\sum_{t\in\mathcal{I}_g}\min\!\left(\rho_{g,t}(\theta)\hat{A}_{g,t},
\operatorname{clip}(\rho_{g,t}(\theta),1-c_-,1+c_+)\hat{A}_{g,t}\right)$.
\end{algorithmic}
\end{algorithm}

\subsection{Turn-Level Reward via Temporal-Difference Credit}
\label{sec:method:td}

\paragraph{One-step TD action credit.}
With no intermediate environment reward and discount $1$, the credit assigned to
the transition $S_k\xrightarrow{a_{k+1},\,o_{k+1}}S_{k+1}$ is the one-step TD
change in value,
\begin{equation}
\delta_k \;=\; V(S_{k+1})-V(S_k)
\;=\;\log\frac{d_k}{d_{k+1}},
\qquad k=0,\dots,T-1 .
\label{eq:td-delta}
\end{equation}
This is positive when the action and returned observation make the gold answer
more likely, equals $\log 2$ when the transition halves the remaining gap, is $0$
when the gap is unchanged, and is negative when the transition moves away from the
answer. Because the credits telescope,
\[
\sum_{k=0}^{T-1}\delta_k \;=\; V(S_T)-V(S_0)
\;=\;\log\frac{-\bar{\ell}_0+\epsilon}{-\bar{\ell}_T+\epsilon},
\]
so total credit depends only on the endpoints: redundant intermediate
transitions cannot inflate it, and the agent is not rewarded for padding a
trajectory.
Appendix~\ref{app:logratio-comparison} compares this log-ratio credit with raw and
linearized alternatives.

\paragraph{$K$-step turn credit.}
The one-step TD credit above can miss delayed tool effects: for example, a \texttt{browser.search}
may only surface candidate links, while the answer likelihood jumps after a
later \texttt{browser.open} exposes decisive evidence.
We therefore use a truncated $K$-step TD backup with $K\ge 1$: each turn receives
discounted future TD progress from the current transition and the next few
transitions in the same trajectory. For rollout $g$ and turn $k$, let
\begin{equation}
h_{g,k}=\min(k+K-1,\,T_g-1)
\label{eq:kstep-end}
\end{equation}
be the last transition included in the backup window. The local progress credit
assigned to the current turn is
\begin{equation}
c^{(K)}_{g,k}
=\frac{1}{Z_{g,k}}
\sum_{u=k}^{h_{g,k}}\gamma_{\mathrm{td}}^{\,u-k}\delta_{g,u},
\qquad
Z_{g,k}=\sum_{u=k}^{h_{g,k}}\gamma_{\mathrm{td}}^{\,u-k},
\label{eq:kstep-credit}
\end{equation}
where $\gamma_{\mathrm{td}}$ discounts delayed evidence and $Z_{g,k}$ is the
$K$-step normalization factor used in our reported runs. In ablations, we use
$K=0$ only as a shorthand for disabling the dense TD backup entirely, not as an
input to Eq.~\ref{eq:kstep-credit}. When the backup window reaches the end of the
trajectory, we additionally anchor the last turns to the verifiable outcome
signal:
\begin{equation}
r^{\mathrm{turn}}_{g,k}
=c^{(K)}_{g,k}
+\mathbf{1}\!\left[h_{g,k}=T_g-1\right]
\lambda_{\mathrm{term}}\,\gamma_{\mathrm{td}}^{\,T_g-k}\,A^{\mathrm{out}}_g .
\label{eq:turn-raw}
\end{equation}
Here $A^{\mathrm{out}}_g$ is the standard GRPO group-relative advantage computed
from the terminal reward, and $\lambda_{\mathrm{term}}$ controls the strength of
the terminal anchor. The exponent $T_g-k$ treats final-answer generation as one
additional transition after the last tool observation, so the last tool turn is
discounted by one step before receiving terminal-outcome fill. This form makes
the two sources of credit explicit: $c^{(K)}_{g,k}$ captures local
answer-likelihood progress, while the terminal fill only affects turns whose
look-ahead window reaches the final outcome. The endpoint telescoping property
above applies exactly to the one-step component $\delta_k$; the $K$-step backup
and terminal fill intentionally trade this exact endpoint-only form for delayed
credit propagation and outcome anchoring.

\paragraph{Joint optimization with outcome reward.}
\label{sec:method:advantage}
We do not group-normalize the turn values. Instead, the TD-derived turn credit is
used directly as a trajectory-local auxiliary signal and is jointly optimized
with the GRPO outcome reward. For the outcome component, we use the standard
group-relative normalization
\[
A^{\mathrm{out}}_g=
\begin{cases}
(R_g-\bar R)/\sigma_R, & \sigma_R>0,\\
0, & \sigma_R=0,
\end{cases}
\quad
\bar R=G^{-1}\sum_{h=1}^{G}R_h,
\quad
\sigma_R=\sqrt{G^{-1}\sum_{h=1}^{G}(R_h-\bar R)^2}.
\]
For each token generated as part of tool-interaction action $a_{g,k+1}$, we set
$\mathrm{turn}(t)=k$; final-answer tokens use the outcome component with the
answer-tail weighting described in Appendix~\ref{app:training-hyperparameters}.
The mixed per-token advantage for tool-interaction tokens is
\begin{equation}
\hat{A}_{g,t}
=\alpha_{\mathrm{out}}\,A^{\mathrm{out}}_g
+\alpha_{\mathrm{turn}}\,r^{\mathrm{turn}}_{g,\,\mathrm{turn}(t)},
\label{eq:combined}
\end{equation}
where $\alpha_{\mathrm{out}},\alpha_{\mathrm{turn}}\ge 0$ control the relative
strength of terminal correctness and turn-level credit. Let $\mathcal{I}_g$ be
the assistant-token indices used in the policy-gradient loss for rollout $g$, and let
$\rho_{g,t}(\theta)=\pi_\theta(a_{g,t}\mid s_{g,t})/
\pi_{\mathrm{old}}(a_{g,t}\mid s_{g,t})$, where $\pi_{\mathrm{old}}$ is the
frozen behavior-policy snapshot that generated the current rollout batch. We
optimize the clipped GRPO objective
\begin{equation}
\mathcal{J}_{\mathrm{TRACE}}(\theta)
=\frac{1}{G}\sum_{g=1}^{G}\frac{1}{|\mathcal{I}_g|}
\sum_{t\in\mathcal{I}_g}
\min\!\left(
\rho_{g,t}(\theta)\hat{A}_{g,t},
\operatorname{clip}\!\left(\rho_{g,t}(\theta),1-c_-,1+c_+\right)\hat{A}_{g,t}
\right),
\label{eq:trace-grpo}
\end{equation}
which keeps the standard GRPO group-relative outcome signal while adding dense
turn-level credit without normalizing the turn values across the prompt group.

%% file: sections/4_experiments.tex
\section{Experiments}
\label{sec:experiments}

\subsection{Experimental Setup}

\paragraph{\textbf{Training datasets.}}
A central question in this work is whether dense turn-level credit can improve
long-horizon tool use when the only verifiable supervision is final-answer
correctness. To examine this question in a controlled setting, we train agents
on synthetic multi-document search tasks built over the offline corpus released
by OpenResearcher~\citep{li2026openresearcher}. Standard
multi-hop QA benchmarks
are often too short for this purpose: many examples can be solved with one
search and one document open, leaving little room to study delayed credit from
later evidence gathering. Our training data instead requires chained retrieval
over multiple irreplaceable evidence documents, so early queries,
document-selection decisions, and in-document localization steps can all affect
the final answer. Appendix~\ref{app:data-synthesis} describes the data
generation, filtering, and verification pipeline.

\paragraph{\textbf{Agentic harness.}}
The policy is trained in a ReAct-style reasoning-and-acting harness
\citep{yao2023react}, where each assistant turn contains private reasoning and
either one browser action or a final answer. The closed-corpus harness serves
documents from the OpenResearcher offline corpus through a FAISS retrieval index
built with Qwen3-Embedding-8B~\citep{zhang2025qwen3embedding}. The browser
interface exposes three actions, \texttt{browser.search}, \texttt{browser.open},
and \texttt{browser.find}; the final response must be emitted inside
\texttt{<answer>} tags. The outcome reward checks normalized exact-match answer
correctness, with a small format component for well-formed answer blocks.
Appendix~\ref{app:search-agent-interface} gives the full tool schema, rollout
stopping rules, masking convention, and prompt template.

\paragraph{\textbf{Models.}}
To evaluate whether the same credit-assignment signal works across model scale,
we instantiate the search policy with two Qwen3 thinking backbones:
Qwen3-4B-Thinking-2507 and Qwen3-30B-A3B-Thinking-2507~\citep{yang2025qwen3}.
The 4B model is our
primary controlled setting for comparing optimization objectives and ablations,
while the 30B-A3B model tests whether \method{} transfers to a larger agentic
backbone. Unless otherwise stated, both models use the same browser action
space, rollout protocol, terminal reward, training data, and evaluation
interface. Importantly, all controlled RL runs start directly from the base
search policy without a cold-start supervised fine-tuning stage, a mid-training
stage, or training on live-web data.

\paragraph{\textbf{Training protocol.}}
We optimize each student with Adam at a constant learning rate of $10^{-6}$,
using a global batch size of $128$ and $8$ rollouts per prompt, with up to $60$ tool
turns per trajectory. \method{} computes turn credit with the log-ratio
formulation in Eq.~\ref{eq:td-delta}, using training gap offset
$\epsilon_{\mathrm{train}}=10^{-1}$. The look-ahead advantage in
Eq.~\ref{eq:turn-raw} uses horizon $K=3$ and discount $\gamma_{\mathrm{td}}=0.8$,
with terminal-outcome fill enabled at scale $2.0$. The final
per-token advantage in Eq.~\ref{eq:combined}
mixes outcome and turn signals with $\alpha_{\mathrm{out}}=1.0$ and
$\alpha_{\mathrm{turn}}=0.2$; the token-level distillation term is disabled in
this work. Tokens in the final answer span are weighted by the group-mean rule
at scale $0.05$. Appendix~\ref{app:training-hyperparameters} lists the full
script-level training, rollout, and turn-credit hyperparameters.

\paragraph{\textbf{Baselines.}}
We compare \method{} with both external deep-research agents and controlled RL
baselines. The external agents include ASearcher-QwQ-32B~\citep{gao2025beyondten},
WebDancer-32B~\citep{wu2025webdancer}, CutBill-30B-A3B~\citep{wu2025cutbill},
and TongyiDS-30B-A3B~\citep{tongyi2025deepresearch}; these systems provide
strong reference points for long-horizon search, although they may use different
data, model backbones, training pipelines, and benchmark harnesses, so we treat
their reported numbers as non-controlled reference points. The controlled baselines are
initialized from the same Qwen3 backbones as \method{}. Base denotes the
untrained search policy. GRPO~\citep{guo2025deepseekr1} uses only the terminal
outcome reward, GSPO~\citep{zheng2025gspo} replaces token-level importance
ratios with sequence-level ratios, and GiGRPO~\citep{feng2025gigpo} applies
group-in-group policy optimization for agent training. For all controlled
baselines, we keep the browser action space, rollout protocol, training data,
terminal reward, and evaluation interface fixed, so performance differences
primarily reflect the optimization objective and credit-assignment signal.

\paragraph{\textbf{Evaluation.}}
We evaluate deep-research ability in both closed-web and open-web settings. The
closed-web benchmark is BrowseComp-Plus~\citep{chen2025browsecompplus}, for
which we build an offline search engine from the officially released corpus
using a Qwen3-Embedding-8B~\citep{zhang2025qwen3embedding} FAISS index. This
setting measures long-horizon reasoning under a reproducible retrieval
environment that matches the training interface. To test transfer beyond the
closed training corpus, we also evaluate on three open-web benchmarks:
BrowseComp~\citep{wei2025browsecomp}, GAIA~\citep{mialon2023gaia}, and
xbench-DeepSearch~\citep{chen2025xbench}, where retrieval is served by the
Serper API~\citep{serper2026}. Among them, xbench-DeepSearch is a Chinese QA
benchmark, allowing us to test whether a training recipe learned in the closed
search setting transfers across languages as well as across retrieval
environments. Across all evaluations, rollouts stop when the model emits a
final \texttt{<answer>} span, reaches the tool-turn limit, or hits a context or
generation limit. Unless otherwise stated, controlled ablations are single
training runs, so small differences should be read as directional rather than as
variance-adjusted conclusions.

\subsection{Main Results}

\begin{table}[t]
\centering
\begin{tabular}{@{}lccccc@{}}
\toprule
Model & BrowseComp-Plus & BrowseComp & GAIA & xbench-DeepSearch & Avg \\
\midrule
\multicolumn{6}{@{}l}{\textit{Deep-research agents}} \\
ASearcher-QwQ-32B & 28.9 & 5.2 & 52.8 & 42.0 & 32.2 \\
WebDancer-32B & 32.4 & 3.8 & 51.5 & 39.0 & 31.7 \\
CutBill-30B-A3B & 35.1 & 13.4 & 46.6 & 43.0 & 34.5 \\
TongyiDS-30B-A3B & 44.4 & 43.4 & 70.9 & 75.0 & 58.4 \\
\addlinespace[0.25em]
\midrule
\multicolumn{6}{@{}l}{\textit{Qwen3-4B-Thinking-2507}} \\
Base & 7.2 & 3.3 & 24.2 & 19.0 & 13.4 \\
GRPO & 30.0 & 5.1 & 38.8 & 44.0 & 29.5 \\
GSPO & 29.7 & 5.4 & 36.7 & 41.0 & 28.2 \\
GiGRPO & 27.7 & 4.4 & 37.9 & 36.0 & 26.5 \\
\rowcolor{gray!15}
\method{} & 35.6 & 6.7 & 44.6 & 49.0 & 34.0 \\
\addlinespace[0.25em]
\midrule
\multicolumn{6}{@{}l}{\textit{Qwen3-30B-A3B-Thinking-2507}} \\
Base & 8.4 & 4.4 & 34.1 & 20.0 & 16.7 \\
GRPO & 36.4 & 10.8 & 45.6 & 37.0 & 32.5 \\
GSPO & 39.7 & 11.8 & 46.6 & 35.0 & 33.3 \\
GiGRPO & 33.0 & 10.1 & 44.7 & 31.0 & 29.7 \\
\rowcolor{gray!15}
\method{} & 42.6 & 12.9 & 52.0 & 45.0 & 38.1 \\
\bottomrule
\end{tabular}%
\caption{\textbf{Performance comparison on long-horizon deep-research benchmarks.}
BrowseComp-Plus is evaluated in the closed-web setting, while BrowseComp, GAIA,
and xbench-DeepSearch are evaluated with open-web retrieval. Avg reports the unweighted
average over all four benchmarks. The top block lists external deep-research
agents for reference, while the Qwen3 blocks compare controlled RL variants
initialized from the same backbone and trained with the same browser interface,
data, terminal reward, and evaluation protocol. Highlighted rows denote
\method{}.}
\label{tab:main_results}
\vspace{-5pt}
\end{table}
\paragraph{\method{} remarkably improves base-model tool use and transfers beyond the training corpus.}
Table~\ref{tab:main_results} shows that \method{} substantially improves
base-model tool use while retaining generalization beyond the training
environment. The most direct evidence comes from the closed-web BrowseComp-Plus
setting, where \method{} raises Qwen3-4B from $7.2$ to $35.6$ and
Qwen3-30B-A3B from $8.4$ to $42.6$ using only RL with outcome-level and
turn-level rewards. This is a strong gain for a minimal recipe: the models are
not first adapted with cold-start SFT, agentic mid-training, or live-web data,
so the improvement suggests that the base backbones already contain useful
reasoning and tool-use primitives, but need a denser credit signal to organize
them into effective long-horizon behavior. The external-agent comparisons give
the result additional context. Although those systems may use different data and
training pipelines, the 4B \method{} agent already surpasses several larger
deep-research agents on BrowseComp-Plus, while the 30B-A3B agent approaches the
strong TongyiDS-30B-A3B result. The more important controlled test is whether
this behavior transfers beyond the closed training corpus: despite being trained
in a closed synthetic search environment, \method{} improves both backbones on
open-web benchmarks served by an external search API; with Qwen3-30B-A3B, it
reaches $12.9$ on BrowseComp, $52.0$ on GAIA, and $45.0$ on the Chinese QA
benchmark xbench-DeepSearch. This pattern suggests that \method{} is not merely
fitting the closed retrieval corpus within our controlled setup; it teaches a
more general interaction strategy for searching, reading observations, and
refining later actions across different corpora and languages, though it does not
close the gap to the strongest external deep-research system in
Table~\ref{tab:main_results}.

\paragraph{\method{} outperforms other RL methods across model scales.}
The controlled RL comparison clarifies why the gains come from credit
assignment rather than from a change in model, data, or environment. All RL
variants use the same backbone, browser interface, training data, terminal
reward, and evaluation protocol; what differs is how the policy-gradient signal
is constructed. Against outcome-only GRPO, \method{} asks the cleanest question:
if the final correctness reward is unchanged, does adding turn-level TD credit
help? The answer is yes on both scales, with the 4B four-benchmark average
rising to $34.0$ from GRPO's $29.5$ and the 30B-A3B average rising to $38.1$
from $32.5$. This
indicates that terminal rewards alone do not expose enough information about
which parts of a long trajectory should be reinforced. GSPO and GiGRPO address a
different aspect of RL optimization by changing sequence- or group-level
estimation, but they still operate primarily at the trajectory level: they can
compare rollouts, yet they do not assign credit to the intermediate interaction
that made later success more likely. This distinction explains why \method{} is
especially beneficial on BrowseComp-Plus and xbench-DeepSearch, where success depends on a
chain of dependent environment interactions rather than a single decisive final
answer. By keeping outcome correctness as the global anchor and adding
temporally aligned progress credit within the trajectory, \method{} supplies the
kind of supervision that long-horizon agent training is missing.
\subsection{Learning Dynamics of \method{}}
\begin{figure}[t]
\centering
\includegraphics[width=\linewidth]{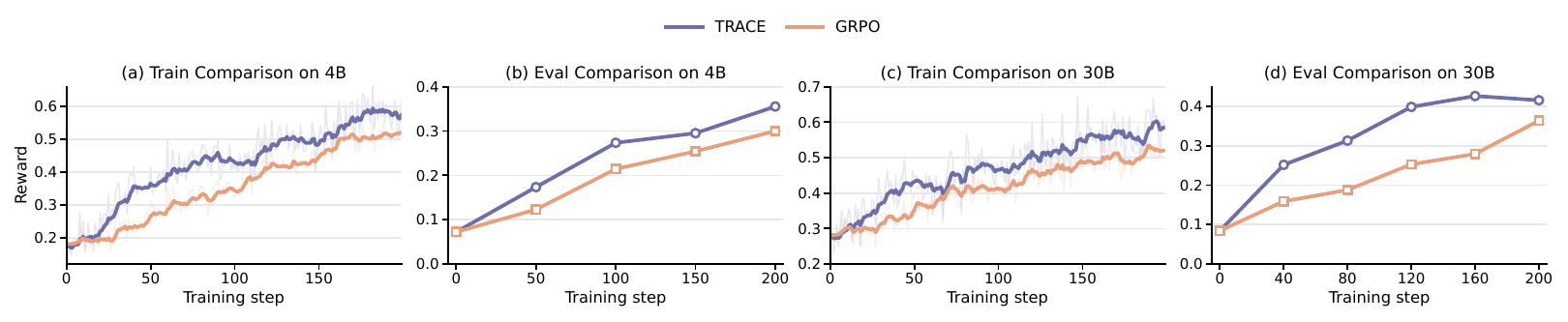}
\caption{\textbf{Learning dynamics on BrowseComp-Plus.} (a) Training reward
on Qwen3-4B. (b) Evaluation accuracy on Qwen3-4B. (c) Training reward on
Qwen3-30B-A3B. (d) Evaluation accuracy on Qwen3-30B-A3B.}
\label{fig:reward_comparison}
\vspace{-0.5em}
\end{figure}

\paragraph{\method{} makes the model converge faster and to a better policy.}
Figure~\ref{fig:reward_comparison} compares the learning dynamics of \method{}
and outcome-reward baselines across four panels. The first visible trend is that
all methods improve as training proceeds, but the \method{} curves start rising
earlier. The second trend is that the slope of \method{} is steeper in the early
stage, indicating faster reward acquisition rather than only a late-stage
separation. The third trend is that the gap does not disappear after the initial
rise: \method{} stabilizes at a higher plateau, suggesting that the faster
optimization also leads to a better learned policy. Finally, the same ordering
appears in held-out accuracy on BrowseComp-Plus with Qwen3-30B-A3B, where the
$160$-step \method{} checkpoint already exceeds the $200$-step outcome-reward
baseline. These observations suggest that the reward-curve improvement reflects
earlier acquisition of transferable long-horizon search behavior. A likely
reason is that outcome-only training assigns one terminal signal to the whole
trajectory, so early failed rollouts can update useful intermediate progress
together with the final mistake. \method{} reduces this temporal mismatch by
assigning turn-level TD credit when an interaction makes the gold answer more
predictable under the reference model, while still keeping the outcome reward in
the objective to enforce final correctness.

\paragraph{\method{} scales up interaction length faster in the beginning stage.}
\begin{wrapfigure}{r}{0.5\linewidth}
\vspace{-0.8em}
\centering
\includegraphics[width=\linewidth]{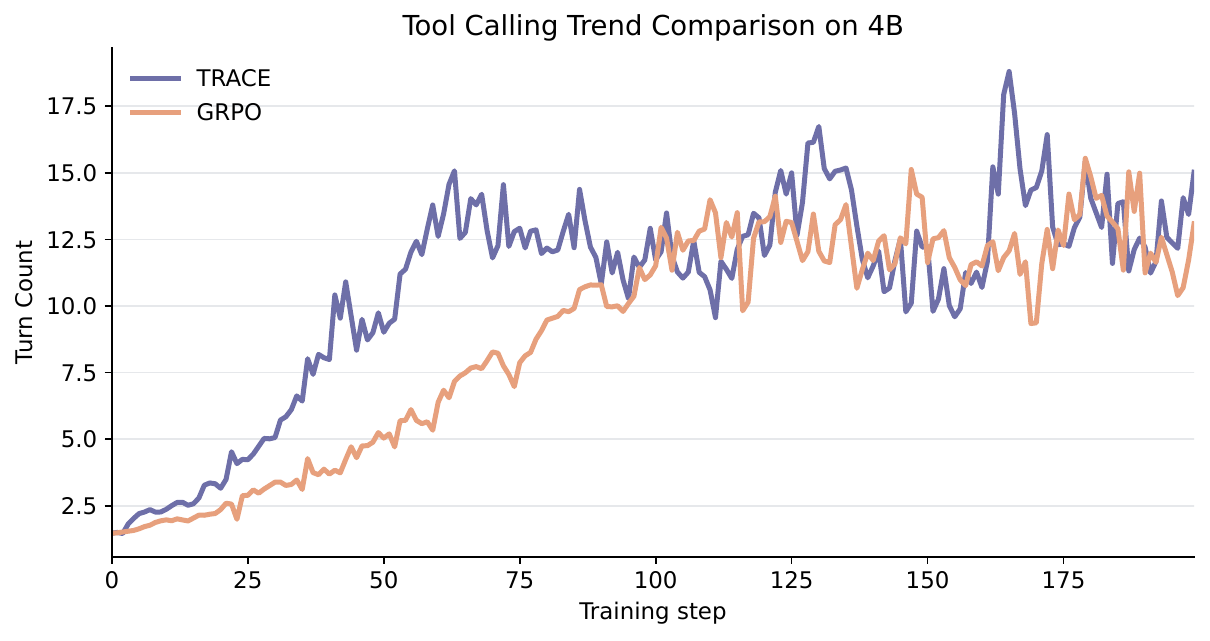}
\caption{Trajectory-scale dynamics on BrowseComp-Plus with Qwen3-4B. The
\method{} curve increases earlier and faster than GRPO, indicating that the two
training objectives expand interaction length at different rates during the
early training stage.}
\label{fig:turn4b_trace_vs_grpo}
\vspace{-1.0em}
\end{wrapfigure}
Figure~\ref{fig:turn4b_trace_vs_grpo} shows a clear difference in trajectory
scale between \method{} and GRPO on the Qwen3-4B backbone: the \method{} curve
starts to rise earlier and grows faster, while the GRPO curve increases more
slowly. This suggests that the objectives differ in how they encourage longer
environment interaction during early training. With outcome-only GRPO, longer
early trajectories often still fail at the final answer, so the whole rollout
receives low or negative relative advantage and the objective cannot separate
unhelpful extra interaction from intermediate progress. \method{} provides a
more local signal: when an interaction and its observation increase answer
readiness, the turn-level TD credit can reward that progress before the final
answer is correct. This gives a smoother incentive for scaling interaction
length, while the outcome reward keeps optimization tied to final correctness.
\subsection{Ablation Study}

\begin{figure}[t]
\centering
\includegraphics[width=\linewidth]{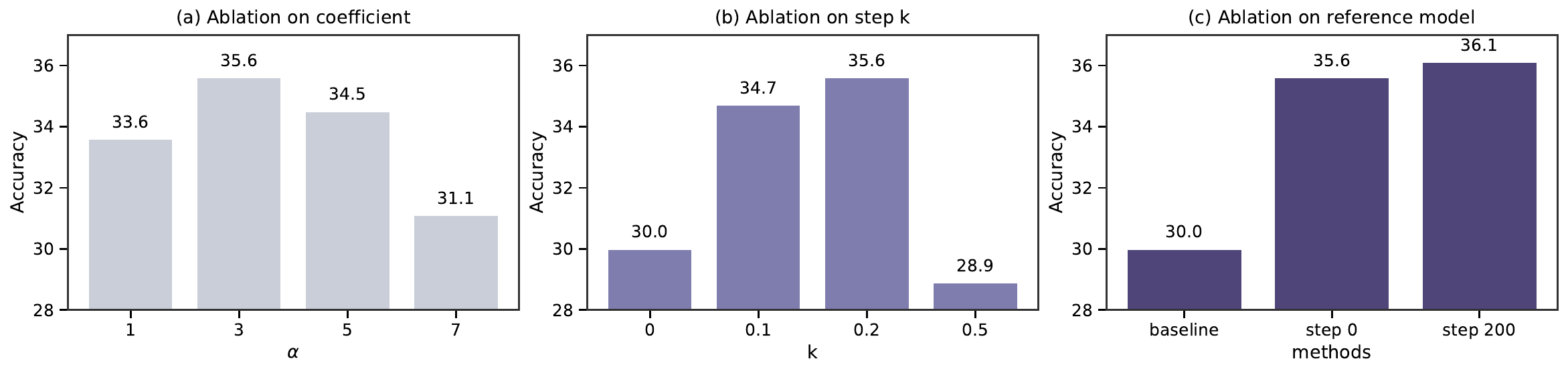}
\caption{\textbf{Ablations on BrowseComp-Plus with Qwen3-4B.} (a) Turn-level
reward coefficient. (b) TD look-ahead horizon $K$. (c) Reference checkpoint for
answer-readiness scoring.}
\label{fig:ablation_study}
\vspace{-0.5em}
\end{figure}

\paragraph{Log-ratio credit assignment is more effective in this run.}
\begin{wraptable}{r}{0.42\linewidth}
\vspace{-0.8em}
\centering
\small
\setlength{\tabcolsep}{4pt}
\begin{tabular}{@{}lc@{}}
\toprule
Method & BC-Plus \\
\midrule
GRPO & 30.0 \\
\quad + raw delta & 32.4 \\
\quad + remaining gap & 34.6 \\
\rowcolor{gray!15}
\quad + log-ratio & 35.5 \\
\bottomrule
\end{tabular}
\caption{Credit-assignment ablation on Qwen3-4B.}
\label{tab:credit_assignment_ablation}
\vspace{-1.0em}
\end{wraptable}
We ablate the turn-level credit format on BrowseComp-Plus while keeping the
same Qwen3-4B training setup. Starting from outcome-only GRPO ($30.0$), adding
dense transition rewards based on the raw log-probability delta improves the
score to $32.4$, and normalizing by the remaining answer-likelihood gap further
improves it to $34.6$. The proposed log-ratio TD credit achieves the best score
in this run ($35.5$), suggesting that relative gap closure provides a more effective credit
signal than absolute likelihood changes. This supports the design in
Eq.~\ref{eq:td-delta}: the log-ratio form better normalizes credit across
states with different confidence levels and preserves a telescoping structure
that discourages redundant trajectory extension.

\paragraph{Ablation on turn-level reward coefficient.}
Figure~\ref{fig:ablation_study}(a) shows that the weight assigned to the dense
turn-level signal must be chosen moderately. A small coefficient underuses the
extra supervision from prefix-level progress, while an overly large coefficient
can make local reference-model readiness dominate the final correctness signal.
In the coefficient sweep, performance rises from $33.6$ to $35.6$ as the
coefficient increases from $1$ to $3$, but then drops to $34.5$ and $31.1$ at
larger values. This pattern supports the use of turn-level reward as an
auxiliary credit signal rather than a replacement for the outcome reward.

\paragraph{Ablation on TD look-ahead horizon $K$.}
Figure~\ref{fig:ablation_study}(b) varies the TD look-ahead horizon $K$,
which controls the strength of the propagated transition-level signal. The
$K=0$ ablation denotes disabling the dense TD backup entirely rather than
passing $K=0$ into Eq.~\ref{eq:kstep-credit}; it obtains only
$30.0$ accuracy, close to outcome-only GRPO, because early search and
document-selection actions receive little useful credit. Moderate propagation
substantially improves performance, reaching $34.7$ and $35.6$, which indicates
that delayed evidence from later \texttt{open} and \texttt{find} operations
should be assigned back to the preceding retrieval decisions. However, the
largest setting reduces accuracy to $28.9$, suggesting that overemphasizing the
propagated signal introduces noise from loosely related later turns and weakens
the temporal locality of the reward.

\paragraph{Ablation on reference model for probability estimation.}
Finally, Figure~\ref{fig:ablation_study}(c) studies which reference checkpoint
should be used to compute the gold-answer prefix probabilities. The baseline
without reference-model progress scoring reaches $30.0$, while using the
initialization checkpoint at step $0$
raises accuracy to $35.6$. Replacing it with a slightly updated checkpoint at
step $200$ obtains a similar score of $36.1$ in this run. The small gap between the two
reference checkpoints suggests that \method{} does not rely on a specially
strong or heavily tuned reference model. Instead, the reference model mainly
serves as a stable value-estimation anchor: it provides prefix-level
answer-readiness scores from which relative TD progress can be computed. In
practice, the base checkpoint is already sufficient for this role, which makes
the method easier to apply and reduces dependence on selecting an advanced
teacher or a carefully adapted scoring model.

%% file: sections/5_relatedwork.tex
\section{Related Work}
\label{sec:related-work}

\paragraph{Post-training of LLMs for reasoning and agentic tasks.}
Post-training has become a central recipe for turning pretrained language models
into capable problem solvers. For reasoning tasks, reinforcement learning with
verifiable rewards has shown that outcome supervision can substantially improve
mathematical and symbolic problem solving when correctness can be checked
automatically \citep{shao2024deepseekmath,guo2025deepseekr1}. These methods
typically optimize a trajectory-level signal with clipped policy-gradient
updates \citep{schulman2017ppo}: a complete solution receives
positive or negative feedback according to whether the final answer is correct.
This simple regime is effective for short-horizon reasoning, but becomes less
informative as decisions grow into multi-hop or browsing trajectories
\citep{yang2018hotpotqa,wei2025browsecomp,chen2025browsecompplus}.
LLM post-training has also been extended from single-response reasoning to
agentic settings in which models interact with tools, environments, users, or
external feedback over multiple turns. Early systems such as WebGPT and ReAct
introduced the interleaving of language-model reasoning, actions, and
observations \citep{nakano2021webgpt,yao2023react}. Subsequent work improves
tool use through self-supervised tool-call prediction, instruction tuning on
tool-use traces, and agent-specific tuning corpora
\citep{schick2023toolformer,qin2023toolllm,zeng2023agenttuning}, while related
approaches use feedback across trials to revise plans, memory, or behavior
\citep{shinn2023reflexion}. This agentic post-training landscape now covers a
wide range of tasks beyond search. Representative settings include shopping and
web navigation
\citep{yao2022webshop,deng2023mind2web,zhou2024webarena,koh2024visualwebarena,drouin2024workarena,pan2024webcanvas},
API and tool use \citep{schick2023toolformer,qin2023toolllm,li2023apibank},
software engineering agents and related coding systems,
real-computer and mobile-device control
\citep{xie2024osworld,rawles2025androidworld,zhang2023appagent}, and multi-turn
user-interactive tool use
\citep{zhao2025muarl,chai2025rlfactory,xi2025agentgymrl,zhang2025agentrl}.
Within software engineering, representative work spans role-based multi-agent
development and interactive code execution
\citep{hong2024metagpt,qian2024chatdev,yang2023intercode,wang2024codeact};
repository-level evaluation, planning, and repair
\citep{jimenez2024swebench,yang2024sweagent,zhang2024codeagent,
bairi2024codeplan,zhang2024autocoderover,bouzenia2025repairagent,
wang2025openhands,xia2024agentless}; and executable training environments,
scalable task generation, and RL or inference-scaling recipes
\citep{pan2025swegym,yang2025swesmith,wei2025swerl,jain2025r2egym,
wang2025swedev,mu2023llm}.
Search and deep-research agents form one important branch of this broader
agentic post-training trend, with recent work applying outcome-based RL,
synthetic trajectory construction, and tool-use training to search-augmented
reasoning and deep browsing
\citep{lisearcho1,chen2025research,jin2025searchr1,song2025r1searcher,guo2025zerosearch,mu2025offline,li2025webthinker,wu2025webdancer,li2025websailor,zheng2025deepresearcher,li2026openresearcher,wang2026sparse,gao2025beyondten}.
Recent surveys and position papers summarize these developments under deep
search, tool-use agents, and agentic deep research
\citep{xi2025deepsurvey,zhang2025agenticdeepresearch}. Our work is in this
general post-training regime: rather than designing a new agent interface or
adding a stronger teacher, we study how to make the training signal more
informative inside a multi-turn trajectory.

\paragraph{Credit assignment for long-horizon language agents.}
Credit assignment is a long-standing challenge in reinforcement learning when
the reward is delayed until the end of an extended trajectory
\citep{sutton2018reinforcement}. Temporal-difference learning addresses this
problem by estimating how each transition changes the value of the current state
\citep{sutton1988learning}. Potential-based reward shaping similarly uses
differences in a state potential to provide denser feedback while preserving the
underlying task objective under standard assumptions \citep{ng1999policy}, and
return-decomposition methods such as RUDDER redistribute delayed rewards to the
decisions that make the final outcome predictable \citep{arjona2019rudder}. In
language-model reasoning, process supervision and process reward models provide
fine-grained feedback by labeling or modeling intermediate reasoning steps
\citep{uesato2022process,lightman2023verify, li2025pqm}. These approaches motivate dense
supervision, but they often require a learned critic, human or synthetic step
labels, process-reward annotation, Monte Carlo continuations, or task-specific
reward features. Recent work has begun to adapt fine-grained credit assignment to
language-model policies. For single-response reasoning, \citet{setlur2024rewarding} formalize a
process reward as the change in the probability of eventually reaching a correct
answer, \citet{wang2024mathshepherd} estimate per-step values from Monte Carlo
rollouts, and \citet{yuan2024free} derive an implicit process reward from the
log-likelihood ratio between a policy and a fixed reference model. For
multi-turn agents, recent work studies stepwise rewards, hindsight credit,
memory-aware credit assignment, search-agent RL, and training infrastructure for
long-horizon tool use
\citep{feng2025gigpo,wang2025spa,dong2025arpo,wang2025stepsearch,zhao2025rsearch,wang2026igpo,tan2026hindsightcredit,yan2026memoryr2,qu2025latentreward,dong2025toolstar,jin2025empiricalsearchrl,xi2025agentgymrl,chai2025rlfactory,zhao2025muarl,zhang2025agentrl}.
\method{} shares the goal of assigning credit below the
trajectory level, but avoids training an additional critic or progress estimator,
does not require Monte Carlo continuations, and does not depend on revisiting identical states. At a high level, \method{} complements outcome-based post-training with turn-level feedback, allowing long-horizon agents to learn from useful intermediate progress rather than only from final success or
failure.

%% file: sections/6_limitations.tex
\section{Limitations}
\label{sec:limitations}

Our empirical validation focuses on long-horizon search agents whose final
answers are short and can be compared against a known ground truth. This setting
matches the core design of \method{}: the frozen reference model estimates the
value of a trajectory prefix by measuring how much more predictable the gold
answer becomes after each tool interaction. For tasks with compact answers, this
provides a simple and stable proxy for progress. However, the same value
estimation strategy may be less direct for agentic tasks whose correct outputs
are long, structured, or open-ended. For example, a code agent may need to
produce a multi-file patch rather than a short answer string, and an open-ended
assistant may need to satisfy underspecified user preferences rather than match a
single gold response. In such settings, it is unclear whether gold-output
log-probability under a frozen reference model is still a reliable state-value
proxy.

This limitation does not affect the main claim that turn-level credit can reduce
the sparsity of outcome-only agentic RL, but it does bound the current scope of
the method. Extending \method{} to richer agent tasks may require alternative
state-value targets, such as execution-based progress signals for coding,
structured task specifications, or decomposed verifiable subgoals. We leave the
design of reliable credit estimators for long-output and open-ended agents to
future work.

%% file: sections/6_conclusion.tex
\section{Conclusion}
\label{sec:conclusion}

We presented \method{}, a critic-free method for assigning turn-level credit in
long-horizon agentic RL. At tool-call boundaries, a frozen reference model maps
gold-answer log-probabilities to log-ratio state values, whose TD changes provide
local rewards. Combining this signal with the GRPO outcome advantage preserves
final-answer correctness as the training anchor without step labels, a strong
LLM judge, or a trained critic or process reward model. \method{} outperforms outcome-only GRPO on all four benchmarks at both model
scales, raising the four-benchmark average from $29.5$ to $34.0$ for Qwen3-4B
and from $32.5$ to $38.1$ for Qwen3-30B-A3B. On BrowseComp-Plus, it improves the
base policies from $7.2$ to $35.6$ and from $8.4$ to $42.6$, with gains
transferring to open-web evaluation. These results establish frozen-reference
turn credit as an effective signal for long-horizon search with compact,
verifiable answers; long, structured, or open-ended outputs will require
alternative state-value estimators.

\section*{Acknowledgment}
We gratefully acknowledge Xuhui Huang for his valuable feedback. The work is supported in part by the Research Forward Fund from the University of Wisconsin-Madison Office of the Vice Chancellor for Research, AFOSR Young Investigator Program under award number FA9550-23-1-0184, National Science Foundation under awards IIS-2237037 and IIS-2331669, Office of Naval Research, Schmidt Sciences Foundation, Open Philanthropy (now Coefficient Giving), and Alfred P. Sloan Fellowship.

%% file: sections/7_appendix.tex
\newpage
\section{Appendix}
\label{sec:appendix}

\subsection{Training Hyperparameters}
\label{app:training-hyperparameters}

Tables~\ref{tab:training-hparams} and~\ref{tab:turn-credit-hparams} summarize
the script-level hyperparameters used for the Qwen3-30B-A3B-Thinking-2507
Search-R1 training run. The shared launch script keeps turn-level scoring
disabled by default for a single-node colocated layout unless a remote
reference-model scoring endpoint is provided; \method{} runs enable this path and use the settings in
Table~\ref{tab:turn-credit-hparams}.

\begin{center}
\refstepcounter{table}\label{tab:training-hparams}
\small
\setlength{\tabcolsep}{4pt}
\begin{tabular}{@{}p{0.18\linewidth}p{0.33\linewidth}p{0.41\linewidth}@{}}
\csname toprule\endcsname
{\bfseries Category} & {\bfseries Hyperparameter} & {\bfseries Value} \\
\midrule
\multirow{14}{*}{Rollout}
	& Rollout engine & SGLang \\
	& GPUs per rollout engine & $1$ \\
	& Running requests & $64$ \\
	& Rollout batch size & $64$ \\
	& Samples per prompt & $8$ \\
	& Sampling temperature & $1.0$ \\
	& Sampling top-$p$ & $1.0$ \\
	& Max trajectory length & $48{,}000$ tokens \\
	& Max response tokens & $4096$ tokens \\
	& Max tool turns & $60$ \\
	& Max observation tokens & $2048$ tokens \\
	& Rollout timeout & $240$s \\
\midrule
\multirow{4}{*}{Evaluation}
	& Eval samples per prompt & $1$ \\
	& Eval max response length & $8192$ tokens \\
	& Eval max context length & $72{,}000$ tokens \\
	& Eval max tool turns & $80$ tool turns \\
\midrule
\multirow{14}{*}{Optimization}
	& Global batch size & $128$ \\
	& Optimizer & Adam \\
	& Learning rate schedule & Constant \\
	& Learning rate & $1\times 10^{-6}$ \\
	& Weight decay & $0.01$ \\
	& Adam $\beta_1$ & $0.9$ \\
	& Adam $\beta_2$ & $0.98$ \\
	& Optimizer offload & CPU offload with overlapped D2H/H2D transfer \\
	& KL loss & $0$ \\
	& Entropy coefficient & $0.00$ \\
	& Clip ratio lower bound & $0.20$ \\
	& Clip ratio upper bound & $0.28$ \\
	& Loss accounting & Per-token loss \\
	& TIS & Enabled \\
\bottomrule
\end{tabular}
\par\vspace{0.5ex}
\parbox{0.96\linewidth}{\small Table~\thetable: Core training, rollout,
and optimization hyperparameters for the Qwen3-30B-A3B-Thinking-2507
Search-R1 training run.}
\end{center}

\begin{center}
\refstepcounter{table}\label{tab:turn-credit-hparams}
\small
\setlength{\tabcolsep}{4pt}
\begin{tabular}{@{}p{0.20\linewidth}p{0.36\linewidth}p{0.36\linewidth}@{}}
\csname toprule\endcsname
{\bfseries Category} & {\bfseries Hyperparameter} & {\bfseries Value} \\
\midrule
\multirow{3}{*}{Outcome reward}
	& Outcome reward coefficient & $1.0$ \\
	& Format score & $0.1$ \\
	& Zero-advantage handling & Return zero when the group advantage is zero \\
\midrule
\multirow{19}{*}{Turn credit}
	& Turn-reward scoring & Enabled for \method{} runs with a remote reference-model scoring endpoint \\
	& Turn-reward weight & $0.2$ \\
	& Training gap offset $\epsilon_{\mathrm{train}}$ & $0.1$ \\
	& Ground-truth field & \texttt{ground\_truth} \\
	& Advantage mode & \texttt{turn\_advantages} \\
	& Remaining-space transform & Enabled \\
	& Transform mode & \texttt{log\_ratio} \\
	& Legacy transform stabilizer & $0.001$ \\
	& TD look-ahead horizon & $K=3$ \\
	& TD discount & $\gamma_{\mathrm{td}}=0.8$ \\
	& $K$-step normalization & Enabled \\
	& Terminal-outcome fill & Enabled \\
	& Terminal-outcome scale & $2.0$ \\
	& Advantage normalization & \texttt{none} \\
	& Post-group normalization & False \\
	& Advantage clipping & $0.0$ \\
\bottomrule
\end{tabular}
\par\vspace{0.5ex}
\parbox{0.96\linewidth}{\small Table~\thetable: \method{}-specific reward and
turn-credit hyperparameters from the Search-R1 launch script. The training gap
offset is the $\epsilon$ used in $d_k=-\bar{\ell}_k+\epsilon$; the legacy
transform stabilizer is a separate script parameter and is not used as the
log-ratio gap offset in the reported \method{} runs.}
\end{center}

\subsection{Comparison with Raw Log-Probability Differences}
\label{app:logratio-comparison}

This section expands on the choice of the log-ratio state value in
Eq.~\ref{eq:value} and the resulting TD action reward in
Eq.~\ref{eq:td-delta}. For a fixed rollout, write $\bar{\ell}_k$ for the
raw average gold-answer log-probability and $d_k=-\bar{\ell}_k+\epsilon$ for
the remaining answer-likelihood gap. The raw log-probability difference
$\Delta^{\ell}_k=\bar{\ell}_{k+1}-\bar{\ell}_k$ telescopes to
$\bar{\ell}_T-\bar{\ell}_0$ and therefore preserves the ordering of two rollouts
from the same prompt when they share the same initial state. However, raw
differences measure absolute improvement in log-probability, not the fraction
of the remaining gap that a turn closes. The linear remaining-gap score
$\widetilde{\Delta}^{\mathrm{lin}}_k=(\bar{\ell}_{k+1}-\bar{\ell}_k)/d_k$ and
the TD reward from the log-ratio value,
$C_k=V_{k+1}-V_k=\log(d_k/d_{k+1})$, both correct this scale issue by rewarding
relative gap reduction, but only the log-ratio value also retains an
endpoint-only telescoping form.

\begin{center}
\refstepcounter{table}\label{tab:delta-comparison}
\small
\begin{tabular}{@{}p{0.22\linewidth}p{0.22\linewidth}p{0.24\linewidth}p{0.22\linewidth}@{}}
\toprule
Property & Raw delta $\Delta^{\ell}_k$ & Linear remaining gap $\widetilde{\Delta}^{\mathrm{lin}}_k$ & Log-ratio TD reward $C_k$ \\
\midrule
Definition & $\bar{\ell}_{k+1}-\bar{\ell}_k$ & $(\bar{\ell}_{k+1}-\bar{\ell}_k)/d_k$ & $V_{k+1}-V_k=\log(d_k/d_{k+1})$ \\
Telescoping & $\bar{\ell}_T-\bar{\ell}_0$ & No endpoint-only form & $V_T-V_0=\log(d_0/d_T)$ \\
Order preservation & Yes, for shared $\bar{\ell}_0$ & Not guaranteed & Yes, for shared $\bar{\ell}_0$ \\
Per-step unit & Absolute improvement & Relative gap reduction & Relative gap reduction \\
Large remaining gap & Uniform scaling & Downweights early large-gap steps & Downweights early large-gap steps \\
Small remaining gap & Uniform scaling & Strongly amplifies near-terminal steps & Moderately amplifies near-terminal steps \\
Numerical behavior & Stable & Can spike when $d_k$ is small & Finite with offset and logarithmic growth \\
\bottomrule
\end{tabular}
\par\vspace{0.5ex}
\parbox{0.96\linewidth}{\small Table~\thetable: Comparison of raw
log-probability differences, a linear remaining-gap normalization, and the TD
reward from the log-ratio value used by \method{}. The log-ratio value is the
only relative-gap formulation whose cumulative action reward depends only on the
initial and final states.}
\end{center}

An offline diagnostic on a held-out BrowseComp-Plus evaluation trace further
supports this choice. We compared the cumulative score from each formulation
against the final reference-model log-probability and the binary outcome reward over
$830$ rollouts containing $3742$ tool turns, using a separate diagnostic offset
$\epsilon_{\mathrm{diag}}=10^{-3}$ for the two remaining-gap normalizations. The log-ratio score has the strongest
correlation with both endpoint quality and final correctness, while also giving
the best pairwise ranking accuracy among rollouts with similar initial raw
log-probabilities.

\begin{center}
\refstepcounter{table}\label{tab:delta-evidence}
\small
\begin{tabular}{@{}lrrr@{}}
\toprule
Metric & Raw & Linear & Log-ratio \\
\midrule
Correlation with final $\bar{\ell}_T$ & $0.425$ & $0.721$ & $\mathbf{0.751}$ \\
Correlation with positive outcome reward & $0.603$ & $0.680$ & $\mathbf{0.713}$ \\
Pairwise ranking accuracy, $|\Delta \bar{\ell}_0|\le 0.2$ & $97.34\%$ & $93.13\%$ & $\mathbf{98.24\%}$ \\
\bottomrule
\end{tabular}
\par\vspace{0.5ex}
\parbox{0.96\linewidth}{\small Table~\thetable: Offline comparison of
transition-credit formulations on a held-out evaluation trace. Scores are
computed by summing turn-level credits within each rollout before comparing to
the final reference-model score or outcome reward.}
\end{center}

The difference is visible at the turn level. Consider two transitions with
nearly identical raw gains: one moves from $\bar{\ell}=-5.1187$ to
$\bar{\ell}=-1.5712$ ($\Delta^{\ell}=3.5475$), while another moves from
$\bar{\ell}=-10.6570$ to $\bar{\ell}=-7.1061$ ($\Delta^{\ell}=3.5509$). Raw
delta treats them as essentially equal. In contrast, the corresponding
log-ratio TD rewards are $1.1806$ and $0.4052$, respectively, because the first
transition closes a much larger fraction of its remaining gap. This behavior is
desirable for long-horizon search: decisive evidence localization near the
answer should receive more credit than an equally sized absolute improvement
that still leaves the rollout far from predicting the gold answer.

\input{sections/8_search_agent}
\input{sections/9_data_synthesis}

%% file: sections/8_search_agent.tex
\subsection{Search-Agent Task and Interface}
\label{app:search-agent-interface}

We instantiate \method{} on a deliberately simple closed-corpus
search task. Given a question $x$, the agent follows an interleaved
reasoning-and-acting pattern \citep{yao2023react}: each assistant turn contains
private reasoning and either one browser tool call or a final answer. If the
turn contains a valid tool call, the environment executes it against the
retrieval corpus and appends the resulting observation before the next assistant
turn. Keeping this interface small helps isolate the effect of the training
signal from improvements due to a more specialized planner or tool stack.

Rollouts stop when the model emits a final \texttt{<answer>} span, reaches the
tool-turn limit, or hits a context or generation limit. Training loss is applied
only to assistant-generated tokens, while tool observations are masked. The
outcome reward checks whether the normalized final answer exactly matches the
gold answer, with a small format component for well-formed answers. We also
record lightweight trajectory diagnostics, including tool-call counts, repeated
queries, reopened documents, repeated in-document searches, navigation errors,
and answer-format validity.

The environment exposes three browsing tools over the retrieval corpus.
Table~\ref{tab:tool-schema} summarizes the tool schema used in all experiments.

\begin{center}
\refstepcounter{table}\label{tab:tool-schema}
\small
\begin{tabular}{lll p{0.39\linewidth}}
\toprule
Tool & Argument & Type/default & Description \\
\midrule
\texttt{browser.search} & \texttt{query} & string, required & Search query string. \\
 & \texttt{topn} & integer, default $10$ & Number of top results to return. \\
\midrule
\texttt{browser.open} & \texttt{id} & integer, default $-1$ & Result index from the most recent search, using zero-based indexing; $-1$ rereads the current page. \\
 & \texttt{loc} & integer, default $-1$ & Starting line number; $-1$ uses the beginning of the page or a relevant passage. \\
 & \texttt{num\_lines} & integer, default $-1$ & Number of lines to display; $-1$ uses the environment default. \\
\midrule
\texttt{browser.find} & \texttt{pattern} & string, required & Exact text pattern to search for in the currently open document. \\
 & \texttt{cursor} & integer, default $-1$ & Page cursor to search in; $-1$ searches the current page. \\
\bottomrule
\end{tabular}
\par\vspace{0.5ex}
\parbox{0.96\linewidth}{\small Table~\thetable: Browser tool schema exposed to
the search agent. Search returns titles, URLs, and summaries for the top
retrieved documents; open displays document content from the closed corpus; find
returns matching line numbers in the currently open document.}
\end{center}

The assistant's reasoning, tool call, and final-answer formatting tokens are
part of the assistant turn for training purposes, while tool observations are
environment outputs and remain masked from the policy-gradient loss.

%% file: sections/9_data_synthesis.tex
\subsection{Synthetic Training Data}
\label{app:data-synthesis}

Training a search agent with verifiable reinforcement learning requires
questions whose rewards remain informative beyond the single-hop retrieval
regime. If the training distribution is dominated by direct lookup questions,
the agent quickly saturates: once it can issue one or two effective searches,
additional training provides little pressure to learn deeper information
gathering or cross-document synthesis. We therefore construct a synthetic
training set of hard multi-document identification questions designed to
require chained search, reasoning over at least two irreplaceable source
documents, and resistance to trivial answer lookup.

\subsubsection{Generation Pipeline}
\label{app:data-synthesis:pipeline}

The pipeline starts from a document corpus, stored as parquet files or a
Hugging Face dataset, together with a related-document index in JSONL format.
For each anchor document, we retrieve the top-$K_{\mathrm{rel}}$ related documents whose
similarity score exceeds a minimum threshold and use this set as the source
context for question generation. This related-document context increases the
probability that a valid question can connect facts across documents while
still requiring the model to identify which documents are necessary.

For every anchor document, we run a best-of-$B$ generation procedure with
rejection sampling. Each independent candidate is generated by an LLM prompted
to produce a hard question, its answer, and an explicit reasoning chain over
the provided sources. If a candidate fails verification, the rejection reason
is fed back to the generator for up to a fixed number of regeneration attempts.
Candidates that pass the initial checks are then sent to an independent answer
verification stage: a separate LLM call receives only the question and source
documents, not the proposed answer, and must derive the answer from scratch.
We discard the example unless the independently derived answer matches the
proposed answer under fuzzy matching. Among the remaining candidates, we keep
the highest-scoring example and write it to the training JSONL.

\subsubsection{Question Templates}
\label{app:data-synthesis:templates}

The generator is constrained to produce questions from a small set of templates
that naturally require cross-document reasoning. Table~\ref{tab:question-types}
summarizes the templates used in our data synthesis pipeline.

\begin{center}
\refstepcounter{table}\label{tab:question-types}
\small
\begin{tabular}{@{}p{0.31\linewidth}p{0.61\linewidth}@{}}
\toprule
Type & Description \\
\midrule
\texttt{BRIDGE\_ENTITY} & Identify a masked entity, then retrieve one of its attributes from another document. \\
\texttt{INTERSECTION} & Find the unique entity satisfying independent conditions from different documents. \\
\texttt{COUNTING\_FILTERED} & Count members of a set defined by cross-document criteria. \\
\texttt{COMPARATIVE} & Compare obfuscated entities using facts distributed across documents. \\
\texttt{REVERSE\_LOOKUP} & Given an obfuscated output, identify its creator or source through another document. \\
\bottomrule
\end{tabular}
\par\vspace{0.5ex}
\parbox{0.96\linewidth}{\small Table~\thetable: Question templates used to
synthesize hard multi-document QA examples.}
\end{center}

To prevent direct keyword lookup, each question applies at least one
obfuscation operation. Time expressions may be blurred into relative or
approximate descriptions; entity names may be masked; names may be replaced by
role descriptions; and quantitative constraints may be converted into
qualitative ones. These transformations force the agent to first resolve what
should be searched for and only then retrieve the evidence needed to answer.

\subsubsection{Verification and Selection}
\label{app:data-synthesis:verification}

Quality control combines programmatic checks with LLM-based verification. The
programmatic check requires the reasoning chain to cite at least two distinct
source URLs. The LLM verifier then evaluates whether the question is actually
hard, grounded, and multi-document. It rejects candidates whose answer appears
directly in the question, whose constraints can be satisfied from a single
source, whose answer is retrievable with a trivial query, or whose reasoning
chain lacks a genuine synthesis step. It also filters unsupported or
hallucinated answers and examples in which one of the cited documents is
redundant.

The final selection score favors candidates that pass both quality and answer
verification, succeed in fewer regeneration attempts, use longer reasoning
chains, cite more distinct documents, and assign primary roles to all cited
documents. It penalizes questions that leak the answer or produce overly long
answers. This scoring rule selects examples that are not merely answerable, but
useful for training long-horizon search behavior: the agent must issue multiple
queries, maintain intermediate entities, and combine evidence from multiple
documents before producing the final answer.

\subsection{Qualitative Analysis of the Turn Credit}
\label{app:qualitative}

Outcome-only GRPO assigns one trajectory-level advantage to every action in a
rollout, obscuring whether an individual turn retrieves decisive evidence,
adds redundant context, or discards an answer already supported by the
transcript. We examine five successful and five failed trajectories from
rollout batches 0140--0149. In each successful case, we highlight the tool
call that closes most of the answer-likelihood gap; in each failed case, the
trajectory first reaches an answer-secured prefix and then loses it after one
diagnostic tool call. Each example reports the complete question and answer,
the one-step credit sequence, and the most diagnostic tool call together with
its credit.

For a trajectory with $T$ tool calls,
\texttt{search\_r1\_turn\_values} contains $T+2$ frozen-reference values:
one before tool use, one after each tool observation, and one after answer
generation. Larger $V$ indicates greater closure of the gold-answer likelihood
gap; we use $V\geq-0.3$ as a diagnostic definition of an
\emph{answer-secured} prefix. Following Eq.~\ref{eq:td-delta},
$\delta_k=V(S_{k+1})-V(S_k)$ is assigned to zero-based tool turn $k$, with the
final difference assigned to answer generation. We mark positive progress in
\textcolor{turnpositive}{green}, harmful transitions in
\textcolor{turnnegative}{red}, and near-zero or mixed credit in
\textcolor{turnneutral}{gray}. Tool calls use the notation
\texttt{browser.tool(argument=value)}; search-relative \texttt{open} IDs are
accompanied by the returned page name.

\subsubsection{Successful Trajectories: Localizing the Turn That Wins}

\begin{tcolorbox}[
	turncreditexample,
	title={Positive turn example P1: A decisive search resolves the distractor}
]
\small
\textbf{Question.}
Starting from the job ad for a major Canadian flag carrier where
line-maintenance applicants are asked to speak English plus German, trace to
that carrier's separate maintenance leadership posting that gives an edge to
bilingual candidates, then compare this carrier against its express-partner
company that says it was recognized as a top small/medium employer around the
mid-2010s. Which company is associated with the mid-2010s employer recognition
(rather than the bilingual-preference maintenance posting)?

\smallskip
\textbf{Answer:} \emph{Air Georgian}. \hfill
\textbf{Final answer:} \finalcorrect

\smallskip
\textbf{Turn-credit distribution.}
\[
\delta=[+6.39,+0.59,+0.10,+0.05].
\]
\textbf{Critical turn.}
\begin{creditturn}{turnpositive}
\textbf{Turn 0.}\enspace \texttt{browser.search(query=question)}\\
\textbf{Turn credit:} $\delta_0=+6.39$.
The salient entity in the question is Air Canada, but the result set surfaces
the Air Georgian careers page and thereby introduces the non-obvious
express-partner answer. The subsequent page open secures the answer, while the
literal \texttt{find(pattern="2015")} adds only $+0.10$ after it is already
secured.
\end{creditturn}
\end{tcolorbox}

\begin{tcolorbox}[
	turncreditexample,
	title={Positive turn example P2: Only the answer-bearing document earns credit}
]
\small
\textbf{Question.}
In the rail-safety material from around the early 2010s, a conference is said
to be opened by a senior infrastructure executive (not named here). Follow
that trail to the police-led plan later described as helping prevent rail
suicides, then move to a later partnership update and identify the person
quoted as managing the suicide-prevention programme for the rail industry. Who
is that person?

\smallskip
\textbf{Answer:} \emph{Ian Stevens}. \hfill
\textbf{Final answer:} \finalcorrect

\smallskip
\textbf{Turn-credit distribution.}
\[
\delta=[+1.63,-1.03,-1.04,+6.05,+0.03].
\]
\textbf{Critical turn.}
\begin{creditturn}{turnpositive}
\textbf{Turn 3.}\enspace \texttt{browser.open(id=2)}
(Samaritans--Network Rail partnership update)\\
\textbf{Turn credit:} $\delta_3=+6.05$.
After two plausible but non-answer-bearing opens, this page names Ian Stevens
and supplies the missing discriminative fact.
\end{creditturn}
\end{tcolorbox}

\begin{tcolorbox}[
	turncreditexample,
	title={Positive turn example P3: A decisive open makes confirmation redundant}
]
\small
\textbf{Question.}
Identify the venue where the Nepali boxer was defeated by the Filipino
Olympian in a unanimous decision, if that Filipino was the same athlete once
knocked out in a world-title rematch by the Thai ex-IBF flyweight champion who
made an impressive Rio start around the mid-2010s.

\smallskip
\textbf{Answer:} \emph{Jakarta Expo Boxing Hall}. \hfill
\textbf{Final answer:} \finalcorrect

\smallskip
\textbf{Turn-credit distribution.}
\[
\delta=[+0.88,+5.86,+0.00,-0.00].
\]
\textbf{Critical turn.}
\begin{creditturn}{turnpositive}
\textbf{Turn 1.}\enspace \texttt{browser.open(id=1)}
(Rappler match report)\\
\textbf{Turn credit:} $\delta_1=+5.86$.
The report supplies ``at the Jakarta Expo Boxing Hall'' and immediately secures
the venue. The subsequent literal \texttt{find} receives $+0.00$, illustrating
that repeated confirmation cannot inflate the telescoping signal.
\end{creditturn}
\end{tcolorbox}

\begin{tcolorbox}[
	turncreditexample,
	title={Positive turn example P4: The search-result snippet secures the answer}
]
\small
\textbf{Question.}
Starting from the maker described as doing Friday evening Arizona ``drops''
and claiming a no-open-edge signature stitch, follow that trail to the
similarly sized product from the black-label line for everyday-carry users,
then compare those with the summer-themed, hand-dyed Atlanta shop item that
comes with a holographic extra. Which brand is the one that explicitly says no
other maker uses its edge/lanyard construction method?

\smallskip
\textbf{Answer:} \emph{Godspeed Tactical}. \hfill
\textbf{Final answer:} \finalcorrect

\smallskip
\textbf{Turn-credit distribution.}
\[
\delta=[+5.48,+0.06,+0.00].
\]
\textbf{Critical turn.}
\begin{creditturn}{turnpositive}
\textbf{Turn 0.}\enspace \texttt{browser.search(query=question)}\\
\textbf{Turn credit:} $\delta_0=+5.48$.
The result snippet exposes the Godspeed Tactical identity and exclusivity
claim before any page is opened. The decisive credit therefore belongs to the
search-result observation rather than the subsequent page read.
\end{creditturn}
\end{tcolorbox}

\begin{tcolorbox}[
	turncreditexample,
	title={Positive turn example P5: A reformulated search rescues a stalled rollout}
]
\small
\textbf{Question.}
In a software podcast episode from around the early 2020s featuring two
engineering guests, an ad segment promotes an open-source time-series database
and references that brand's event series. For the London edition that was
moved online during the pandemic period, what was the earlier company name
(from around the early 2010s) that immediately preceded the host company's
later name change associated with that database?

\smallskip
\textbf{Answer:} \emph{Errplane}. \hfill
\textbf{Final answer:} \finalcorrect

\smallskip
\textbf{Turn-credit distribution.}
\[
\begin{aligned}
\delta={}&[+1.18,-0.12,-0.19,-0.14,\\
&\phantom{[}+0.43,+5.44,+0.02,+0.00].
\end{aligned}
\]
\textbf{Critical turn.}
\begin{creditturn}{turnpositive}
\textbf{Turn 5.}\enspace
\texttt{browser.search(query="influxdata company name change from early 2010s
to later name")}\\
\textbf{Turn credit:} $\delta_5=+5.44$.
After four stalled probes, this reformulated query returns ``Errplane became
InfluxData in 2015,'' localizing nearly all progress in the eight-turn
trajectory to this rescue query.
\end{creditturn}
\end{tcolorbox}

\subsubsection{Negative Trajectories: Localizing the Tool Call That Loses}

Every failure below reaches an answer-secured prefix and then undergoes a
single large negative transition. The examples differ in what triggers the
collapse: a query-echo self-match, an over-interpreted no-match, a match in the
wrong context, an over-literal type comparison, or a phrase copied from the
question rather than read from the page.

\begin{tcolorbox}[
	turncreditexample,
	title={Negative turn example N1: A query-echo self-match discards a secured answer}
]
\small
\textbf{Question.}
In a post from around the mid-2010s by a youth organizer planning an
end-of-season teen church overnight, the writer jokes about a very high-voltage
fence and doubts a loft drop trust exercise. Use that to find another
reflection from roughly the same era where students mention trusting the
rope-holder during a heights course, then compare the fear-focused activity
named there with the similarly fear-themed activity listed in a different
school day-log that also mentions climbing, archery, a blindfold obstacle
course, and a ladder challenge. Which fear-themed activity title comes later
in alphabetical order?

\smallskip
\textbf{Gold answer:} \emph{Leap of Faith}. \quad
\textbf{Model answer:} \emph{loft drop trust exercise}.\\
\textbf{Final answer:} \finalincorrect

\smallskip
\textbf{Turn-credit distribution.}
\[
\delta=[+1.47,+0.13,+1.49,-0.00,-4.83,-0.78].
\]
\textbf{Critical turn.}
\begin{creditturn}{turnnegative}
\textbf{Turn 4.}\enspace
\texttt{browser.find(pattern="loft drop trust exercise", cursor=0)}\\
\textbf{Turn credit:} $\delta_4=-4.83$.
After the gold answer is secured, this call matches the model's earlier query
echo rather than a page-supported title. Accepting its own paraphrase as a
second candidate discards the answer and produces the sharp negative credit.
\end{creditturn}
\end{tcolorbox}

\begin{tcolorbox}[
	turncreditexample,
	title={Negative turn example N2: A no-match becomes a false factual negative}
]
\small
\textbf{Question.}
Starting from the article about a TV doctor's health-content partnership
launched on a global blood-disorder awareness observance in the late 2010s,
use that observance date to find another program episode from around the
mid-2010s featuring three women in the same disease space. Then use the
early-2010s Pittsburgh interview with the longtime local nonprofit executive
(described by role, not name) to filter that trio by shared family name. How
many women in the trio match that surname?

\smallskip
\textbf{Gold answer:} $2$. \quad \textbf{Model answer:} $0$. \hfill
\textbf{Final answer:} \finalincorrect

\smallskip
\textbf{Turn-credit distribution.}
\[
\begin{aligned}
\delta={}&[+2.31,+0.10,-1.32,+0.09,\\
&\phantom{[}+0.35,+0.98,-4.05,-1.09].
\end{aligned}
\]
\textbf{Critical turn.}
\begin{creditturn}{turnnegative}
\textbf{Turn 6.}\enspace \texttt{browser.find(pattern="Bush")}
(returns ``No matches'')\\
\textbf{Turn credit:} $\delta_6=-4.05$.
Although the preceding prefix is answer-secured, the model interprets absence
from an incomplete episode page as evidence that none of the women is a Bush.
The negative credit localizes this invalid factual-negative inference.
\end{creditturn}
\end{tcolorbox}

\begin{tcolorbox}[
	turncreditexample,
	title={Negative turn example N3: A successful find locks onto distractor context}
]
\small
\textbf{Question.}
Starting from a celebrity-photo page from around the mid-2010s that praises two
aristocratic sisters for attending separate events in England, follow to the
article comparing the elder future queen consort's conservative red charity
gown with her younger sibling's more revealing red look earlier that month.
Using those two sisters as the candidate set, then check a later royal-fashion
tag page from the early 2020s and count how many of those candidates are
explicitly mentioned in the sentence about children serving as page attendants
and bridesmaid at the younger sister's wedding.

\smallskip
\textbf{Gold answer:} $2$. \quad \textbf{Model answer:} $1$. \hfill
\textbf{Final answer:} \finalincorrect

\smallskip
\textbf{Turn-credit distribution.}
\[
\begin{aligned}
\delta={}&[+2.19,-0.35,+0.10,-1.03,+0.96,\\
&\phantom{[}+0.50,-3.63,+0.19,-1.22].
\end{aligned}
\]
\textbf{Critical turn.}
\begin{creditturn}{turnnegative}
\textbf{Turn 6.}\enspace
\texttt{browser.find(pattern="Pippa Middleton")}\\
\textbf{Turn credit:} $\delta_6=-3.63$.
The call succeeds, but matches ``as she did for her aunt Pippa Middleton last
year,'' a sentence about the wrong wedding. From a just-secured prefix, the
model anchors on this superficially relevant single-name match and later
answers $1$. Tool success is therefore not equivalent to information progress.
\end{creditturn}
\end{tcolorbox}

\begin{tcolorbox}[
	turncreditexample,
	title={Negative turn example N4: Exact-string matching replaces a type comparison}
]
\small
\textbf{Question.}
Using a post from around the early 2010s by the person who says he ordered
interchangeable raw-brass ring parts as your starting clue, go to a different
instrument-forum thread that references a repair specialist by first name and
mentions that writer considering an instrument with many strings. From that
thread, take the distinct instrument types named there as the candidate set.
Then check a separate dealer page listing ``other instrument offerings'' from
the mid-2000s era and count how many of those candidate types are present
there.

\smallskip
\textbf{Gold answer:} $1$. \quad \textbf{Model answer:} $0$. \hfill
\textbf{Final answer:} \finalincorrect

\smallskip
\textbf{Turn-credit distribution.}
\[
\delta=[+1.42,-0.10,+0.82,+0.08,-3.41,-0.97].
\]
\textbf{Critical turn.}
\begin{creditturn}{turnnegative}
\textbf{Turn 4.}\enspace \texttt{browser.find(pattern="autoharp")}
(returns ``No matches'')\\
\textbf{Turn credit:} $\delta_4=-3.41$.
The dealer page has already secured the answer, but the task requires comparing
the candidates by instrument type. The model instead probes one literal string
and treats its absence as proof that the intersection is empty, discarding the
page-level evidence.
\end{creditturn}
\end{tcolorbox}

\begin{tcolorbox}[
	turncreditexample,
	title={Negative turn example N5: A prompt phrase is mistaken for page text}
]
\small
\textbf{Question.}
Using the workshop notice from around the early 2010s that mentions a big
northern English city and a hotel-branded venue, trace forward to a later
full-day procurement training page to recover a per-attendee cost, then apply
that amount as a strict threshold to the industry-only Canadian pricing
workshops held a few years afterward. How many city sessions in that Canadian
notice have fees greater than that threshold?

\smallskip
\textbf{Gold answer:} $2$. \quad \textbf{Model answer:} $0$. \hfill
\textbf{Final answer:} \finalincorrect

\smallskip
\textbf{Turn-credit distribution.}
\[
\begin{aligned}
\delta={}&[+2.00,+0.43,+0.06,-2.46,\\
&\phantom{[}+0.40,+1.98,-3.48,-1.03].
\end{aligned}
\]
\textbf{Critical turn.}
\begin{creditturn}{turnnegative}
\textbf{Turn 6.}\enspace
\texttt{browser.find(pattern="fees greater than")} (returns ``No matches'')\\
\textbf{Turn credit:} $\delta_6=-3.48$.
The Canadian workshop page has already supplied the city-level fees. The model
nevertheless searches for a comparison phrase copied from the question rather
than comparing the listed numbers, then over-interprets the no-match and
answers $0$.
\end{creditturn}
\end{tcolorbox}
Across the positive cases, nearly all credit lands on the \texttt{search} or
\texttt{open} that first supplies the missing discriminative fact; redundant
probes and post-hoc \texttt{find} calls remain near zero. Across the negative
cases, all five trajectories first reach the answer-secured region, after which
one misleading \texttt{find} receives the dominant penalty. The calls cover
both empty matches that are over-interpreted and successful matches in the
wrong context, showing that tool completion alone does not imply progress.
Outcome-only GRPO cannot express any of these within-trajectory distinctions
because it assigns every turn the same trajectory-level advantage.